# Multi-Labelled Value Networks for Computer Go

Ti-Rong Wu[1], I-Chen Wu[1], *Senior Member, IEEE*, Guan-Wun Chen[1], Ting-han Wei[1], Tung-Yi Lai[1], Hung-Chun Wu[1], Li-Cheng Lan[1]

*Abstract* — This paper proposes a new approach to a novel value network architecture for the game Go, called a multi-labelled (ML) value network. In the ML value network, different values (win rates) are trained simultaneously for different settings of komi, a compensation given to balance the initiative of playing first. The ML value network has three advantages, (a) it outputs values for different komi, (b) it supports dynamic komi, and (c) it lowers the mean squared error (MSE). This paper also proposes a new dynamic komi method to improve game-playing strength.

This paper also performs experiments to demonstrate the merits of the architecture. First, the MSE of the ML value network is generally lower than the value network alone. Second, the program based on the ML value network wins by a rate of 67.6% against the program based on the value network alone. Third, the program with the proposed dynamic komi method significantly improves the playing strength over the baseline that does not use dynamic komi, especially for handicap games. To our knowledge, up to date, no handicap games have been played openly by programs using value networks. This paper provides these programs with a useful approach to playing handicap games.

## I. INTRODUCTION

Go is a two player perfect information game that originated more than 2,500 years ago in China. The game is simple to learn, but difficult to master. Two players, *black* and *white*, alternately place one stone of each player's color on one empty intersection, referred to as *points* in this paper, on a board, usually a 19x19 board for most players. Normally, black plays first. A player can capture opponent stones by surrounding them, or make *territory* by surrounding empty points. The goal of players is to occupy as many points of their own as possible. A game ends when both players can occupy no more territory. Agreement to end the game is often indicated by choosing to pass consecutively during each player's turn.

When a game ends, the *territory difference* is calculated to be black's territory minus white's. To compensate for the initiative black has by playing first, a certain number of points, the so-called *komi*, are added for white (the second player), balancing the game. A game's score is its territory difference minus komi. Black wins if the score is positive, and white wins otherwise. Komi is customarily set to 7.5 according to Chinese rules, and 6.5 in Japanese rules.

Although the rules of Go are simple, its game tree complexity is extremely high, estimated to be $10^{360}$ in [1][40]. It is common for players with different strengths to play $h$-stone handicap games, where the weaker player, usually designated to play as black, is allowed to place $h$ stones[2] first with a komi of 0.5 before white makes the first move. If the strength difference (rank difference) between both players is large, more handicap stones are usually given to the weaker player.

In the past, computer Go was listed as one of the AI grand challenges [16][28]. By 2006, the strengths of computer Go programs were generally below 6 kyu [5][8][14], far away from amateur dan players. In 2006, *Monte Carlo tree search* (MCTS) [6][11][15][23][37] was invented and computer Go programs started making significant progress [4][10][13], roughly up to 6 dan in 2015. In 2016, this grand challenge was achieved by the program AlphaGo [34] when it defeated (4:1) Lee Sedol, a 9 dan grandmaster who had won the most world Go champion titles in the past decade. Many thought at the time there would be a decade or more away from surpassing this milestone.

Up to date, DeepMind, the team behind AlphaGo, had published the techniques and methods of AlphaGo in Nature [34]. AlphaGo was able to surpass experts' expectations by proposing a new method that uses three *deep convolutional neural networks* (DCNNs) [24][25]: a *supervised learning* (SL) policy network [7][9][18][26][38] learning to predict experts' moves from human expert game records, a *reinforcement learning* (RL) policy network [27] improving the SL policy network via self-play, and a *value network* that performs state evaluation based on self-play game simulations. AlphaGo then combined the DCNNs with MCTS for move generation during game play. In MCTS, a fast *rollout policy* was used to compute state evaluations in the past. AlphaGo uses a combination of fast rollout policy with the value network for state evaluation.

In AlphaGo, the RL policy network and value network were trained based on a komi of 7.5 to conform to most computer Go competitions. For the games played by human players on the Go game servers Tygem and Fox, the komi was set to be 6.5. However, since the RL policy network and value network were trained for a komi of 7.5, slightly less accuracy is expected when playing games with a different komi.

A naive solution would be to retrain another set of RL policy network and value network with a new komi when needed. The same could be proposed for handicap games with a komi of 0.5. However, the cost of training each new set of RL policy network and value network is high, due to the finite amount of precious computing resources that are available. Additionally, the playing strength of each new set of networks could be different or lead to inconsistencies.

[1] This work was supported in part by the Ministry of Science and Technology of the Republic of China (Taiwan) under Contracts MOST 104-2221-E-009-127-MY2, 104-2221-E-009-074-MY2 and 105-2218-E-259-001. The authors are with the Department of Computer Science, National Chiao Tung University, Hsinchu 30050, Taiwan. (e-mail of the correspondent: icwu@csie.nctu.edu.tw)

[2] In Go handicap games, the $h$ stones are placed on designated points, except when $h$=1. The details are available at [32].

For this problem, currently, the value networks that were trained based on a komi of 7.5 are often used to play 6.5-komi games. In practice, few games end up in the special category where the outcome would be different between a komi of 6.5 and 7.5. Unfortunately, as unlikely as this may be, it has happened in real games; e.g., the program DeepZenGo's losing match [21] during the World Go Championship [29] against Mi Yuting, one of the top players in the world ranking [12], who, to date, had a highest ranking of being the third strongest player. At the time, one of the authors of DeepZenGo proposed a trick [20] to offset the expected inaccuracies by setting komi to 5.5 during rollout, which, when combined with the evaluation by the value network that was trained with a komi of 7.5, balanced the program to behave as if the komi was 6.5.

To solve this problem elegantly, this paper proposes a new approach to the design of a *multi-labelled* (*ML*) *value network* modified from the value network architecture. The ML value network is able to train the win rates for different komi settings simultaneously. Moreover, we also incorporate *board evaluation* (*BV*) into the ML value network, named *BV-ML value network*. BV can be useful in that it expresses point ownership by player as a probability so that there exists an estimate on each point (or territory) on the board. This estimation ultimately correlates with the multi-labelled values. With these new approaches, the ML value network can not only offer games with different komi, but also support dynamic komi (which was not supported in [34]) to improve the game playing strength. This paper also proposes a new dynamic komi method based on the BV-ML value network to improve the strength significantly, especially for handicap games.

Our proposed approach's advantages are then corroborated by our experiment results, as follows.

1. The *mean squared error* (*MSE*) of the BV-ML value network is generally lower than the value network alone.
2. The program based on the BV-ML value network wins by a win rate of 67.6% against the program based on the value network alone. This shows that the BV-ML value network significantly improves the playing strength.
3. We propose a method to measure the prediction of the final score from ML value networks, and the experimental results show high confidence on the predicated score. This hints that the predicted score can be used for dynamic komi.
4. The proposed dynamic komi method improves the playing strength with the following experimental results. The program with the dynamic komi method wins against a baseline by win rates of 80.0%, 57.2%, 41.6% and 18.4% respectively for 1-stone, 2-stone, 3-stone and 4-stone handicap games. In contrast, the program without the dynamic komi method wins by 74.0%, 50.0%, 30.4% and 10.8%.
5. We analyze the effect of using the BV-ML value network trained from 7.5-komi games to play on 6.5-komi games. The experiments demonstrate that the BV-ML value network slightly improves over the value network without ML.

The remainder of this paper is organized as follows. Section II reviews some past work, such as policy networks, the value network, BV, MCTS and dynamic komi. Section III presents our network architecture and dynamic komi method. Section IV does experiments to analyze and demonstrate the merits of our architecture and method. Finally, Section V draws conclusions.

## II. BACKGROUND

This section reviews the SL/RL policy network and the value network in Subsection II.A, describes the combination of DCNNs and MCTS in Subsection II.B, introduces the board evaluation network in Subsection II.C, and discusses the dynamic komi technique for MCTS [3] in Subsection II.D.

### A. SL and RL Policy Network and Value Network

The goal of the supervised learning (SL) policy network is to predict the possible moves of human experts with a given position. The prediction of human expert moves is useful in selecting moves (or actions) that are very likely to be played, thereby reducing the branching factor significantly. Networks that served action selection were referred to as *policy networks* in AlphaGo's terminology. During the development of AlphaGo, 28.4 million human expert positions and moves were collected as *state-action pairs (s, a)* from the KGS Go Server to train a 13-layer SL policy network, where *s* are positions and *a* are the moves corresponding to *s*. During training, state-action pairs were randomly sampled among these collected pairs; the sampled state-action pair was then used to maximize the likelihood of the expert move given the state *s* using gradient ascent. The training procedure described above followed the equation:

$$\Delta \sigma \propto \frac{\partial log P_\sigma(a|s)}{\partial \sigma}$$

where $\sigma$ are the weights of the SL policy network, $P_\sigma$ is the probability of taking action *a* on state *s*, following the policy network. The SL policy network of AlphaGo had 48 input channels and 192 filters, and was able to achieve a prediction rate of 57.0% with all input features.

To further improve the policy network, AlphaGo trained a policy network through self-play via reinforcement learning (RL) [35][36]. Since the RL policy network was built from the basis of the SL policy network, its structure was identical to the SL policy network, with the RL policy network weights $\rho$ initialized to the same values as the SL policy network. That is, $\rho = \sigma$ initially. The RL training process started by listing a policy pool, with only the SL policy network in the pool initially. In each iteration, a random opponent was chosen from the pool, then 128 games were played. The game results $z_t$ were then used to update the RL network by policy gradient. After 500 iterations, the current policy was added to the opponent pool. The above described process followed the REINFORCE algorithm [41], which is summarized below:

$$\Delta \rho \propto \frac{\partial log P_\rho(a_t|s_t)}{\partial \rho}(z_t - v(s_t))$$

The game results $z_t$ is modified by a baseline value, $v(s_t)$. For the first round, the self-play baseline value was set to zero. On the second round, the value network $v_\theta(s_t)$, which will be described in the next paragraph, was set to be the baseline. By doing so, AlphaGo was able to achieve a slightly better performance. With the RL training complete, AlphaGo was able to win 85% of the games played against Pachi [4], an open-source computer Go program. For reference, with just the SL policy network, it was only able to win 11% against Pachi.

The goal of the value network is to estimate a value (or a win rate) for any given game position. Theoretically, there exists an optimal value function $v^*(s)$ given perfect play. In practical situations, since perfect play is not possible, AlphaGo attempted to arrive at an estimated value function $v^{P_\rho}$. This estimated value function is based on the performing policy using the RL policy network. The value function is approximated with a network $v_\theta(s)$, where $\theta$ are the weights of the network, and $v_\theta(s) \approx v^{P_\rho}(s) \approx v^*(s)$. The value network's structure is similar to the SL policy network, with a small difference in that the value network outputs a single scalar as a prediction of the value of the given position, while the SL policy network outputs a probability distribution. Overfitting is a major concern since the number of human expert game records are limited. To overcome this potential problem, AlphaGo uses the RL policy network to create 30 million new game records (unique game positions) through self-play. Each self-play game was performed by playing the same RL policy network against itself.

### B. MCTS and DCNNs

Monte Carlo Tree Search (MCTS) [6] is a best-first search algorithm on top of a search tree [23][33], using Monte Carlo rollouts to estimate state values. A *tree policy*, traditionally the *upper confidence bounds* (*UCB*) function [2], is used to select positions (states) from the root to one of the leaves. When a leaf position is chosen, one or more children are expanded from the selected leaf position.

AlphaGo used the PUCT algorithm [31] (serving as the tree policy) to select states in a tree, and applied the SL policy network and the value network to MCTS by asynchronous updates. The SL policy network is used in MCTS expansion, where MCTS takes the probability outputs of child states expanded from the network and uses them as the initial values of the PUCT algorithm. The evaluation consists of two parts, the value network output and the Monte Carlo rollout. For the first part, the value network is used in the evaluation of leaf positions $S_L$. For the second part, Monte Carlo rollouts are performed, starting from $S_L$ until the end of the game. Compared with the value network, rollouts can generate estimated values quickly but with less accuracy. During MCTS selection, the value of each position is calculated by a value combining the results of the value network and the rollout with a weighting parameter λ, which was set to 0.5 in AlphaGo. To minimize end-to-end evaluation time, AlphaGo uses a mini-batch size of 1.

### C. Board Evaluation Network

Besides the value network, another neural network with territory called the board evaluation (BV) network was proposed previously [19]. The number of outputs of the BV network is the same as the number of board points. Each output value corresponds with a board point, indicating the probability that the point belongs to black by the endgame (the probability for white is simply the complement of this value). The BV network's output layer uses a sigmoid activation, while the network's objective is to minimize the MSE between predicted probabilities and the actual endgame ownership of each board point (taking the form of one of two possible values). In [19], the BV network uses 8 input channels, 5 layers with 64, 64, 64, 48, 48 filters in each layer, using the Go4Go dataset as training data. To date, none have exploited this idea further except a few discussions given in [30].

### D. Dynamic Komi

In traditional computer Go programs with MCTS, *dynamic komi* [3] is a technique that is widely used to make the program play more aggressively, especially for handicap games. One of the known deficiencies of MCTS-based programs is its inability to handle extremely favorable or unfavorable situations. MCTS works by maximizing the winning probability instead of considering score margins between the players, so when the outcome of the game is already apparent, there is very little incentive to either further strengthen one's advantage when winning or to catch up and minimize one's losses when losing.

There are two methods in [3] that adjust the komi dynamically, providing this incentive to play better in extreme conditions. The first is the so-called *score-based situational* (*SS*) method, which adjusts the komi based on $E[score]$, the expected score of rollouts over a specific amount of Monte Carlo simulations. The new komi is set to $k + E[score]$, where $k$ is the game's komi. In order not to be so aggressive, the new komi can be set to $k + \alpha E[score]$, where $\alpha$ is a value, called the *komi rate*. The komi rate is set to be higher near the beginning of a game, and lower near the end. This is because in the beginning of a game, win rates are less precise, so komi adjustments allow the program to play more proactively even when the program seems to be winning or losing decisively. Near the end of a game, the win rates are more precise, so there is less need to adjust komi significantly and frequently.

Another is the *value-based situational* (*VS*) method, which adjusts the komi dynamically so that over a specific amount of Monte Carlo simulations, the win rate of the root of the tree, $v$, falls into a predefined interval $(l, u)$ (e.g. (45%, 50%) [3]), or at least as close to the interval as possible. Namely, if $v$ is within the interval, the komi is not adjusted. If $v$ is higher than $u$, increase komi by one, and if lower than $l$, decrease by one. In [3], some criterion is also used to avoid oscillation of komi values, especially in the endgame.

For the above two methods, the specific amount of Monte Carlo simulations between each adjustment can be set to either (a) the amount of simulations needed for a single move, or (b) a specific number of rollouts, e.g. 1000 [3]. For the latter, the komi is adjusted faster, but this may incur inconsistent win rates, which are directly impacted by different komi values. Previous research reported to work fine with the latter [3], while Aya, a notable Go program, chose to use the former [42]. In [3], the author also mentioned the possibility of increasing or decreasing by more than one when using the VS method. However, no detailed methods and experiments were reported.

### III. OUR DESIGN

We present our network design in Subsection III.A, and dynamic komi methods in Subsection III.B.

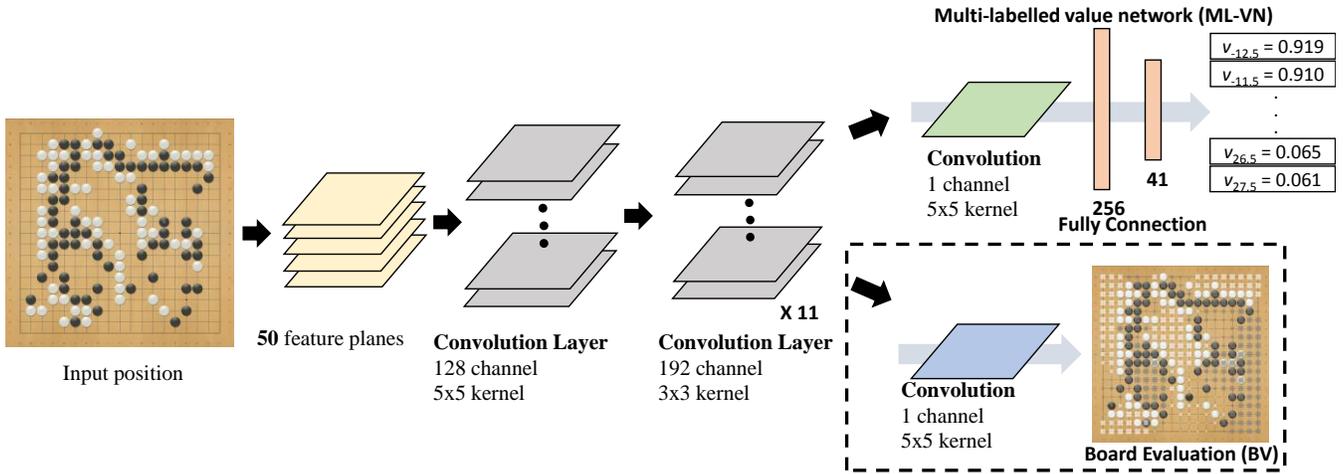

Figure 1. The network architecture of our BV-ML-VN. The network has 13 layers. The input feature planes are calculated from the given position. There are two outputs. One output is the multi-labelled value $v_k$ for a range of komi values $k$, and the other output is the board evaluation which indicates the ownership of each point for the given position.

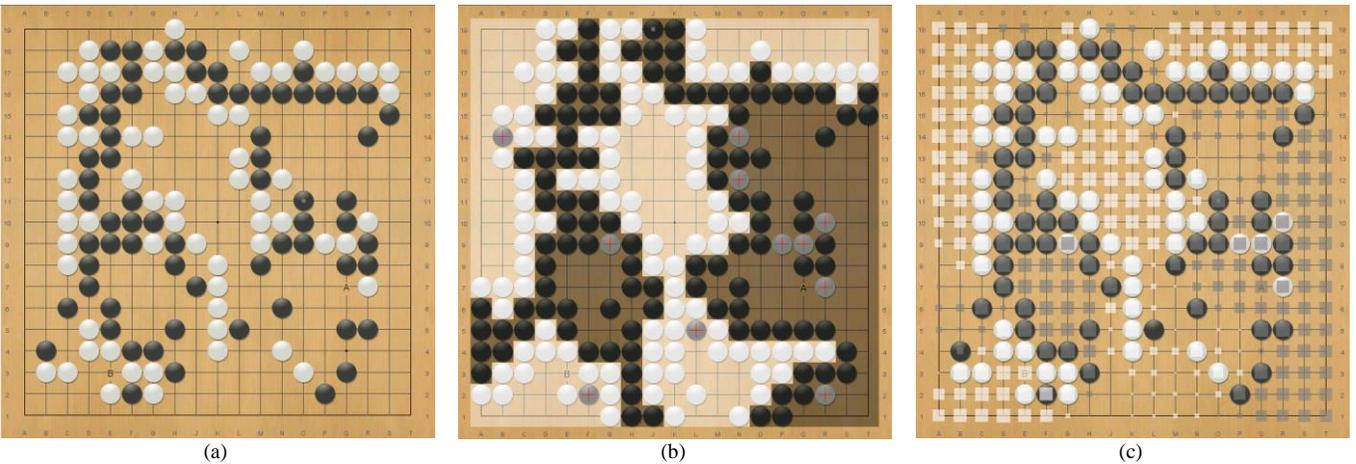

(a)          (b)          (c)

Figure 2. (a) A position of a KGS game for illustration. (b) The end game of the position (a), where the black (white) shadows indicates black's (white's) territory. (c) The ownership output from the inference of BV-ML value nework for position (a), where the sizes of black (white) squares indicate the confidence of ownership by black (white).

## A. Network Design

In general, our SL and RL policy networks follow AlphaGo's with some slight modifications as follows. In addition to the 48 input channels used by AlphaGo, we simply add one more input channel for the $ko^3$ feature, indicating the ko position when there is a ko in current game state. As a side note, we trained the SL policy network with 3-step prediction, which was proposed in [38]. In our training data, we combine samples from the game records played in the KGS Go game server [22] (only for those by 6 dan players or stronger) and the open game records played by professional players collected in GoGoD [17]. For the training results, the SL policy network prediction rate reached 56% for the KGS testing data, and 52% for the GoGoD testing data. For the purposes of this paper, our RL policy network only went through one round of training (no baseline value is used). Without a value network to provide a baseline value, the resulting strength of our policy network won 70% of games against Pachi [4].

After training the SL/RL policy networks, we use our RL policy network to generate 30 million self-play games. Each self-play game ends when the ownerships of all points are determined, instead of ending as soon as one side has won. This is because in addition to evaluating wins/losses, we also need to know the number of winning points so that the training sample contains the win rates for different komi settings, as well as the final territory for BV.

This paper proposes a new design for the value network, named the *multi-labelled (ML) value network*. AlphaGo's value network only has one output indicating the value of the given position, namely, the win rate of the position with a komi of 7.5. In our approach, the value network includes a set of outputs $v_k$, each indicating the value of the position for $k$-komi games. For simplicity of discussion, all win rates are for black in the remainder of this paper. From the rules of the game, the full set of outputs can be $v_{-361.5}$ to $v_{361.5}$. However, for simplicity, we only consider $v_{-12.5}$ to $v_{27.5}$, centered at $v_{7.5}$, in our design. The ML value network, shown in Figure 1 excluding the dashed box, has 13 layers (the same structure as SL/RL policy networks) with the last two layers being fully connected layers, where the first has 256 rectifier units and the second with 41 rectifier units each of which is finally connected to a tanh unit output to one $v_k$ ($v_{-12.5}$ to $v_{27.5}$). Similar to AlphaGo's value network, the ML value network also adds one more input channel to indicate the player color.

---

[3] In Go, "ko" refers to situations where players are able to capture stones alternately, causing board repetition. Playing ko is forbidden so that endless cycles can be avoided [32].

For the training data, we label on output $v_k$ as follows. For each self-play game, first calculate territory difference $n$ at the end of the game. Then, based on the Chinese rule, label 1 (win) on $v_k$ for all $k < n$, and -1 (lose) for all $k > n$. (Note that the draw case $k = n$ is ignored in this paper since the komi is not an integer normally.) For example, if black occupies 7 more points of territory than white, the $k$-komi game is considered a win for all $k < 7$, and a loss for all $k > 7$. Thus, in this case, a 7.5-komi game is a loss, and a 6.5-komi or 0.5-komi game is a win.

| $k$ (komi) | Label on $v_k$ | $v_k$ (win rate) |
|---|---|---|
| -3.5 | 1 | 0.800943 |
| -2.5 | 1 | 0.748812 |
| -1.5 | 1 | 0.748309 |
| -0.5 | 1 | 0.680036 |
| 0.5 | 1 | 0.678897 |
| 1.5 | 1 | 0.599618 |
| 2.5 | 1 | 0.599108 |
| 3.5 | -1 | 0.512413 |
| 4.5 | -1 | 0.511263 |
| 5.5 | -1 | 0.423886 |
| 6.5 | -1 | 0.423626 |
| 7.5 | -1 | 0.339738 |
| 8.5 | -1 | 0.339353 |
| 9.5 | -1 | 0.265258 |
| 10.5 | -1 | 0.264586 |
| 11.5 | -1 | 0.20581 |
| 12.5 | -1 | 0.204716 |

Table 1. The multi-labelled output for different komi values $k$.

Table 1 illustrates the labelling of $v_k$ (listing $v_{-3.5}$ to $v_{12.5}$ only for simplicity) for the position in Figure 2 (a) (above), with the game ending in the position in Figure 2 (b), where black owns 3 more points in territory than white. Thus, for all $k < 3$, the outcome $v_k$ is labeled as 1, and −1 for all $k > 3$. In Table 1, while the second column lists the labelled values $v_k$ as above, the third column lists the value $v_k$ for the inference on the position of Figure 2 (a) by our ML value network. These values $v_k$ decrease as $k$ increases, since the win rate at $k$-komi is no higher than that at $k$+1-komi.

Table 1 also shows an interesting result: every two win rates of the ML value network output, namely $v_{2k-0.5}$ and $v_{2k+0.5}$ for each integer $k$, are very close in value. This phenomenon is because assuming all points are occupied by either black or white when the game ends, the territory difference (based on the Chinese rule) must be an odd number. The only possible case where the territory difference is even is when games end with an odd number of *seki*[4] points. However, our statistics indicate that only 1.5% of our self-play games end this way[5].

Furthermore, we incorporate BV into our value network, for which we refer to as *board evaluation and multi-labelled* (*BV-ML*) *value network* as follows. In the BV-ML value network, additional networks are added into the value network as shown in the dashed box of Figure 1. The network in the dashed box shares the same input channels and 12 layers as the ML value network. From there, the output of the 12th layer is connected to an additional convolution layer that is connected to a sigmoid layer outputting the predicted ownership $O_P$ of each point $P$ on board. A value network with BV but without the multi-labelled component is referred to as a *BV value network*.

Let us illustrate by the position shown in Figure 2 (a). Let the position shown in Figure 2 (b) be the end of the self-play game starting from the position in Figure 2 (a). Consider two points, $A$ at coordinate Q7 occupied by black and $B$ at E3 by white at the end of the game (as shown in Figure 2 (b)). The output $O_A$ is labelled with 1, and $O_B$ is labelled with 0. The ownership outputs from the inference of the BV-ML value network are shown in Figure 2 (c).

We also apply these value networks to our computer Go program "CGI Go Intelligence" (abbr. CGI), placed 6th in the 9th UEC Cup [39]. The MCTS method basically follows AlphaGo's with some slight modifications as follows. Different from AlphaGo using a mini-batch size of 1 to minimize end-to-end evaluation time, we use a larger batch size of 16. The reason is simply to speed up performance by increasing the throughput. In addition, we also proposed and implemented new dynamic komi methods, as described in the next subsection.

*B. Dynamic Komi*

This subsection first modifies both SS and VS methods (described in Subsection II.D) in order to support dynamic komi on the Go programs with value networks, and then also proposes a new dynamic komi method.

For the SS method, dynamic komi is adjusted based on the territory at the root node which can be generated from rollouts, board evaluation of the BV network, or a mix of the two. Let SS-R, SS-B and SS-M represent the SS method using the three kinds of territory generations respectively. In SS-M, the weighting parameter $\lambda = 0.5$ is used, like the way the value is mixed from the rollouts and the value network in [34]. The formula of komi rate is the same as [3].

For the VS method, if the current dynamic komi is $k$, the next dynamic komi is adjusted based on the win rate at the root node, $w_k = (1 - \lambda)r_k + \lambda v_k$, where $r_k$ is the win rate of rollouts and $v_k$ is the value at komi $k$ from the ML value network. Let VS-M represent the above modified VS method. Similarly, the parameter $\lambda$ is also 0.5. Note that we do not consider other values for $\lambda$, namely, where $\lambda = 0$ for rollouts alone and $\lambda = 1$ for the value $v_k$ alone.

This paper also proposes a new approach to supporting dynamic komi on value networks by leveraging multi-labelling. In this approach, all the mixed win rates $w_k = (1 - \lambda)r_k + \lambda v_k$ are maintained. Like the VS method, we want the mixed win rates $w_k$ for the dynamic komi $k$ to fall into a predefined interval $(l, u)$, or at least as close to the interval possible. Let $k0$ be the game's komi. If $w_{k0}$ is within the interval, the next dynamic komi is still $k0$ for the next move. If it is out of the interval, say $w_{k0} > u$ without loss of generality, locate the closest komi $k$ such that $w_k \leq u$. Then, the next dynamic komi is set to $k0 + \alpha(k - k0)$, where $\alpha$ is the komi rate. For example, if $k0$ is 7.5, $w_{10.5}$ is 57% and $w_{11.5}$ is 53%, then the next dynamic komi would then be adjusted to 11.5 assuming $\alpha = 1$ and $(l, u) = (45\%, 55\%)$.

---

[4] In Go, seki refers to local patterns where some points, called seki points in this paper, are adjacent to two live groups of stones of different colors and neither player can fill them without dying [32].

[5] With the Japanese rules, since it is possible to end the game without counting some of the last played moves, the chance of even territory differences are higher. The details on Japanese rules are available at [32].

**procedure ML-BASED DYNAMIC KOMI**
1. *Require:*
2. *$i$: the ordinal number of the current move to play;*
3. *$w_k$: the mixed win rate of the root for all komi k;*
4. *$k0$: the real komi for this game;*
5. *$B$: the total points of the whole board, 361 in $19 \times 19$ Go games;*
6. *$c$ and $s$: parameters to decide different komi rate;*
7. *$u$ and $l$: the contending interval $(u,l)$;*
8. **if** $Value < l$ **then**
9.   Locate a komi $k$ such that $w_k \geq l$ and $w_{k-1} < l$.
10. **else if** $Value > u$ **then**
11.   Locate a komi $k$ such that $w_k \leq u$ and $w_{k-1} > u$.
12. **else**
13.   Locate komi $k$ as the $k0$;
14. **end if**
15. $KomiDiff \leftarrow k - k0$
16. $GamePhase \leftarrow i / B - s$
17. $KomiRate \leftarrow (1 + exp(c \cdot GamePhase))^{-1}$
18. $DynamicKomi \leftarrow Komi + KomiDiff \cdot KomiRate$
**end procedure**

The above new dynamic komi routine, named **ML-BASED DYNAMIC KOMI (ML-DK)**, contains parameters which are listed in Lines 2-7. The target komi is determined in Lines 8-14. The komi rate decreases by a small amount as games progress, according to the formula in Lines 15-17, and the next dynamic komi is calculated based on the komi rate in Line 18. The komi rate depends on parameters, $c$ and $s$, set to 8 and 0.45 respectively in this paper. Figure 3 (below) shows how the komi rate changes as the game progresses.

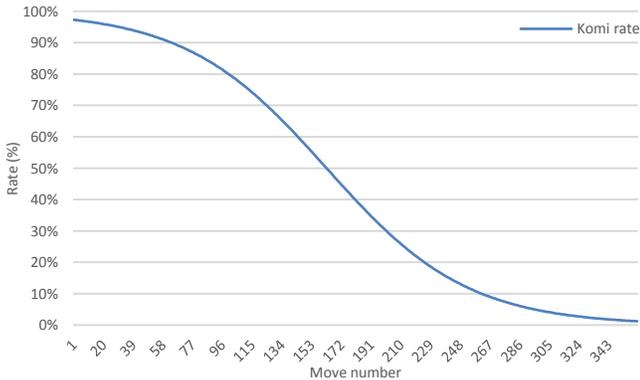

Figure 3. The value for komi rate in different move numbers.

## IV. EXPERIMENTS

In this section, all the experiments are done on machines equipped with four GTX 980Ti GPUs, two Intel Xeon E5-2690s (28 cores total), 2.6 GHz, 128 GB memory, and run on Linux. In Subsection IV.A, we introduce four different network architectures and measure their mean squared errors (MSEs) and playing strengths. In Subsection IV.B, we analyze the MSEs of these networks for games of different komi values to verify the advantage of multi-labelled (ML) value networks. The accuracy of the number of winning points in BV-ML value network are presented and analyzed in Subsection IV.C. Subsection IV.D discusses different dynamic komi techniques in MCTS with different handicap games. Subsection IV.E analyzes the performance difference when using $v_{7.5}$ and $v_{6.5}$ to play 6.5-komi games. Finally, we analyze the correlation of BV and ML value network in Subsection 0.

### A. Different Value Networks

We use our RL policy network (as described in Subsection III.A) to generate 30 million self-play games, which with four GTX 980Ti GPUs takes about 3 weeks.

We implement four different value networks: just the value network, the value network with BV only, the ML value network, and the BV-ML value network. For simplicity, they are denoted respectively by VN, BV-VN, ML-VN and BV-ML-VN. When updating, the loss weight ratio between BV and VN is set to 1:1. Each network has been trained to convergence, which with four GTX 980Ti GPUs takes about 2 weeks each.

To measure the quality of these value networks, we analyze their MSEs like AlphaGo [34] did. Three sets of KGS games were used as our benchmark, where at least one of the players is 6 dan or higher. The three sets are: a set of 0.5-komi games chosen at random from KGS, a set of 6.5-komi games, and a set of 7.5-komi games. Each of the sets listed above consists of 10,000 games. For the $i$-th game in one set, assume that the result of the end game is $z_i$, where $z_i$ is 1 for a black win, and $-1$ otherwise. Let $s_{ij}$ denote the $j$-th position in the $i$-th game, and $v(s_{ij})$ denote the value obtained by inputting the position into the given value network. The MSE($j$), or the MSE at all the $j$-th positions for all $n$ games, where $n = 10,000$ in each set, is calculated by the following formula.

$$\frac{1}{2}\sum_{i=1}^{n}\left(z_i - v(s_{ij})\right)^2$$

For simplicity, all the $j$-th positions where $j \geq 285$ is counted in MSE(285).

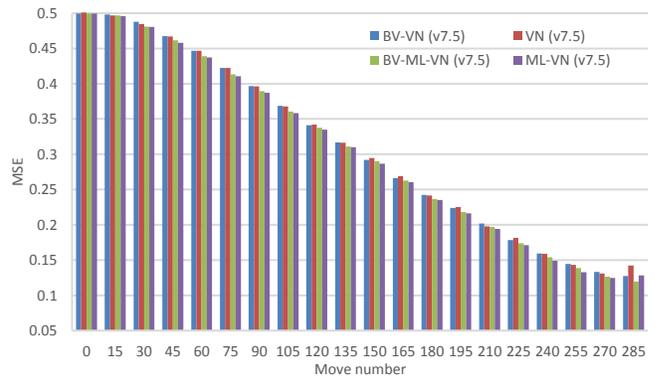

Figure 4. MSE for different value networks in 7.5-komi games.

For fairness, we use the set of 7.5-komi games as testing data, and the MSEs of the four value networks are shown in Figure 4. Since the numbers are very close, we also show the average of these MSEs in Table 2 (below). The results show that both ML-VN and BV-ML-VN have lower MSEs than VN and BV-VN.

However, the MSE of BV-ML-VN is not lower than that of ML-VN. The reason is that the BV output only provides the probability of each point's ownership at the end of the game. It has nothing to do with the win rates, nor the MSE of the win rates.

| Network architecture (abbr.) | Mean squared error (MSE) |
|---|---|
| VN | 0.35388 |
| ML-VN | 0.346082 |
| BV-VN | 0.35366 |
| BV-ML-VN | 0.348138 |

Table 2. The average MSE of different value networks in 7.5-komi games.

| Network | VN | ML-VN | BV-VN | BV-ML-VN |
|---|---|---|---|---|
| VN | - | 39.60% (±4.29%) | 39.40% (±4.29%) | 32.40% (±4.11%) |
| ML-VN | 60.40% (±4.29%) | - | 49.20% (±4.39%) | 47.20% |
| BV-VN | 66.60% (±4.29%) | 50.80% (±4.39%) | - | 47.20% (±4.38%) |
| BV-ML-VN | 67.60% (±4.11%) | 52.80% (±4.38%) | 52.80% (±4.38%) | - |

Table 3. Cross-table of win rates with 95% confidence between different value networks in 7.5-komi games.

We apply the four value networks respectively to our computer Go program (CGI) to compare the strengths among these networks, for which the results are shown in Table 3. For each pair of programs (with different value networks), 500 games are played with 1s/move, while one GPU and six CPU cores are used for each program. The results shows that the programs with BV-ML-VN, BV-VN and ML-VN significantly outperform the program with VN only, and that BV-ML-VN performs the best. Although the BV-VN does not perform better than ML-VN/BV-ML-VN in terms of the MSE, the strength of BV-VN is nearly equal to ML-VN and close to BV-ML-VN. For the above reason, most of experiments in the remaining of this section are based on BV-ML-VN.

### B. Multi-labelled Value Network in Different Komi Games

To measure the network quality in different komi games, we analyze two other data sets, the 6.5-komi and 0.5-komi games. Since the MSE of both VN and BV-VN are nearly the same and the MSE of both ML-VN and BV-ML-VN are nearly the same (also from the previous subsection), we only show the MSE of BV-VN and BV-ML-VN for simplicity.

First, we analyze the MSE of value networks for 6.5-komi games. For BV-VN, since the output only contains one label (trained from 7.5-komi games), it is used to estimate win rates for these games. For BV-ML-VN, we use the value $v_{6.5}$ to estimate win rates of these games, instead. Figure 5 (below) depicts the MSE of BV-VN and BV-ML-VN in 6.5-komi games. The MSE for BV-ML-VN is slightly lower than that for BV-VN, but the difference is not much, since few 6.5-komi games ended with black winning by 0.5 points (usually marked as "B+0.5" in Go game scoring). Note that 6.5-komi games with "B+0.5" are losses if the games were played under a komi of 7.5 points. In our statistics, among all the collected games, only 0.78% 6.5-komi games ends with "B+0.5". Since these are only a very small portion of all games, most programs with value networks directly play all games based on a komi of 7.5, even for games with a komi of 6.5.

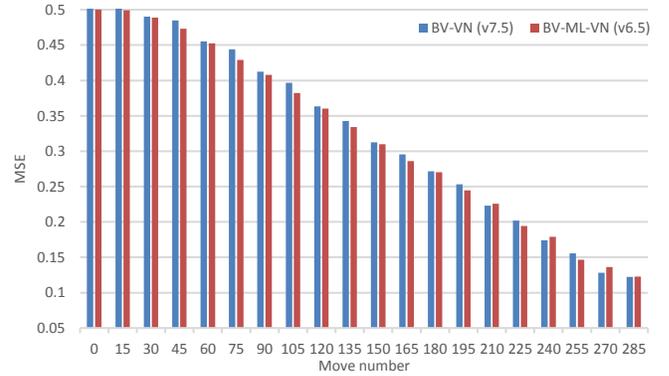

Figure 5. MSE for different value networks in 6.5-komi games.

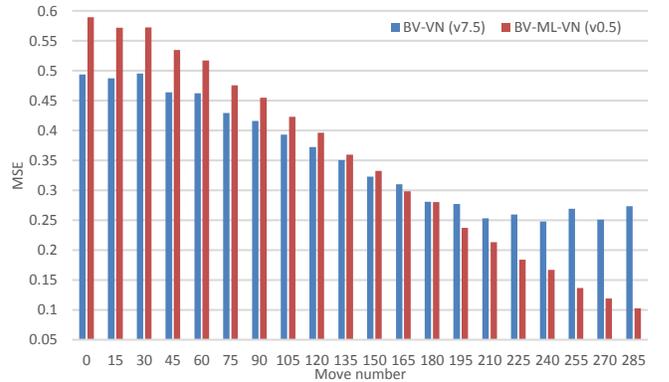

Figure 6. MSE for different value networks in 0.5-komi games.

Next, we analyze the MSE of value networks for 0.5-komi games. For BV-ML-VN, we use the value $v_{0.5}$ to estimate win rates of these games. Figure 6 depicts the MSE of BV-VN and BV-ML-VN in 0.5-komi games.

From the figure, it is interesting that the MSE of BV-ML-VN is much higher than that of BV-VN in the early stage of games, namely, about 0.6 for BV-ML-VN versus 0.5 for BV-VN at the beginning of games. The initial (empty) games are nearly even, namely about 50% for 7.5-komi games, since they are trained from 7.5-komi games (assuming both players have equal strength). Thus, the values in BV-VN and $v_{7.5}$ in BV-ML-VN are about 50%. Based on this, when the komi is changed to 0.5, this implies that the game favors black, and the win rate of black in the beginning should be more than 50%. In reality, we observed a win rate of 63% for black in our BV-ML-VN.

However, the collected 0.5-komi games are usually played by players of different ranks, where white is often one dan higher than black. From these collected games, we found that black only obtained a win rate of 39.88%, which resulted in an even higher MSE. However, after 150 moves or so, the MSE of BV-ML-VN becomes lower than that of BV-VN's. This is because BV-VN always uses a komi of 7.5 to estimate win rates, resulting in a progressively inaccurate evaluation as the game goes on. Statistically, only 5.33% of 0.5-komi games had the results "B+0.5" to "B+6.5"; game results in this range are interesting because they will have different outcomes under these two circumstances. Namely, they are considered winning if they are played with a komi of 0.5, but losing for a komi of 7.5. From the calculation of MSE (as described in Subsection IV.A), an error rate of 5.33% results in larger MSE values as shown in Figure 6. In contrast, BV-ML-VN converges (as VN does for 7.5-komi games), since BV-ML-VN can accurately estimate the situation based on the output for $v_{0.5}$.

*C. The Accuracy of Winning Points in BV-ML-VN*

This subsection analyzes the accuracy of winning points in the BV-ML-VN. Previously, with only one label, the BV-VN only outputs an estimate of the win rate. With multiple labels for different values of komi, BV-ML-VN is instead able to evaluate the win rates for each komi. With this extra information, we can then estimate the point difference between the two players.

For our analysis, we first locate a komi $k$ such that $v_k \geq 50\%$ and $v_{k+1} < 50\%$. Then, we expect black to win if the komi is smaller than and equal to $k$, and lose if the komi is larger than and equal to $k+1$. As mentioned in the komi rules in Section I, territory differences in real games can only be integers. That is, black leads by $k + 0.5$ without komi. For example, in Table 1, $k$ is 4.5, since $v_{4.5} \geq 50\%$, $v_{5.5} < 50\%$, and the BV-ML-VN expects black to lead by 5 points without komi. Since the game has a komi of 7.5, the network predicts "W+2.5" for the position corresponding to the table. The outcome of the real game is "W+4.5" (as shown in Section III.A), i.e., black leads by 3 points without komi. For simplicity of discussion, we say, in the remainder of this subsection, that the network predicts black is leading by 5 points (without komi) and that the real outcome of the game is black leading by 3 points. The difference of the point gap predicted by the network and the actual outcome of the game is called the *prediction distance*. In the above example, the prediction distance is 2.

Now, we want to analyze the accuracy of our value network in the following manner. First, from the set of 7.5-komi KGS games (as described in Subsection IV.A), we choose the games ending with black leading -12 to 28 points (also excluding games that were resolved through resignations), since our BV-ML-VN is trained with multiple labels from $v_{-12.5}$ to $v_{27.5}$.

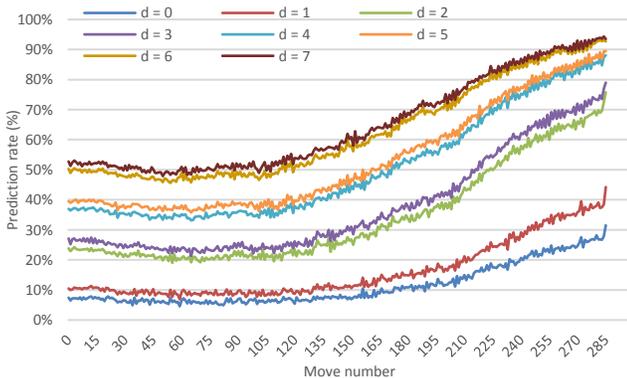

Figure 7. $d$-prediction rates of BV-ML-VN.

A game is said to $d$-predict at move $i$ correctly, if the prediction distance of the $i$-th position of the game is smaller than or equal to $d$. Consequently, $d$-prediction rates at move $i$ are the percentage of games that $d$-predict at move $i$ correctly. Figure 7 shows $d$-prediction rates for $d = 0$ to 7.

For each $d$, the prediction rates generally grows higher as the game progresses. As $d$ increases, the $d$-prediction rates also increase. From the figure, we observe that the lines tend to "pair up", i.e., $d$-prediction rates for $d = 2k, 2k + 1$, tend to be close. The reason is similar to the one given for the paired consecutive win rates $v_{2k}$ and $v_{2k-1}$ in Table 1.

When $d$ is 0, the prediction rates are about 8% at the initial stage, and increase up to 32% near the end of the game. Three factors negatively impact this prediction rate. First, human players may have consistently chosen sub-optimal moves due to miscalculation or decided to play too conservatively to secure victory, resulting in less than optimal values. Second, the network may not be able to predict accurately. Third, KGS uses Japanese rules which may cause a one point difference in some games. When $d$ is 1, the prediction rates are slightly higher about 10% at the initial stage, and increase up to 40% near the end of the game.

When $d$ is 2 or 3, the prediction rates significantly improve, from about 20% in the beginning of the game to 75% near the end. The same can be observed for $d = 4$ or 5, and $d = 6$ or 7. When $d$ is 6 or 7, the prediction rates are about 45% at the initial stage, and grow up to 92% near the end. The analysis shows high confidence for short prediction distance even with the three negative factors pointed out in the previous paragraphs. This finding hints at the accuracy of dynamic komi; more experiments about dynamic komi are described in Subsection IV.D.

*D. Dynamic Komi and Handicap Games*

This subsection analyzes the performances of six different dynamic komi methods, namely, no dynamic komi, SS-R, SS-B, SS-M, VS-M, and ML-DK as described in Subsection III.B. For both VS-M and ML-DK, the win rate interval is (45%, 55%). For SS-* and ML-DK, we choose the setting of komi rates by letting parameters $c = 8$ and $s = 0.45$.

The above dynamic komi methods are incorporated into our Go program CGI with BV-ML-VN. In this experiment, the baseline we choose is the version of our early CGI program which participated in the 9th UEC Cup. For each dynamic komi method, let the program with the method play against the baseline 250 games, each of which is played on $19 \times 19$ boards with one second per move. Since our BV-ML-VN supports $v_{-12.5}$ to $v_{27.5}$ only and one-stone handicap was valued at about 7 points difference according to the experiments in [3], we only analyze 1-stone to 4-stone handicaps, in addition to even games. The komi is set to 0.5 in handicap games, and 7.5 in even games.

| Method | Even Game |
|---|---|
| No Dynamic komi | 80.80% ($\pm 4.89\%$) |
| SS-R | 80.00% ($\pm 4.97\%$) |
| SS-B | 78.00% ($\pm 5.15\%$) |
| SS-M | 79.60% ($\pm 5.01\%$) |
| VS-M | 82.00% ($\pm 4.77\%$) |
| ML-DK | 83.20% ($\pm 4.64\%$) |

Table 4. Performance of dynamic komi in even Games.

Table 4 presents the result for various dynamic komi methods in even games. The result shows that the version with ML-DK outperforms all other versions.

| Method | H1 | H2 | H3 | H4 |
|---|---|---|---|---|
| No dynamic komi | 74.00% (±5.45%) | 50.00% (±6.21%) | 30.40% (±5.71%) | 10.80% (±3.86%) |
| SS-R | 76.00% (±5.30%) | 58.00% (±6.13%) | 38.00% (±6.03%) | 22.00% (±5.15%) |
| SS-B | 77.60% (±5.18%) | 54.00% (±6.19%) | 31.60% (±5.77%) | 20.40% (±5.01%) |
| SS-M | 74.40% (±5.42%) | 49.20% (±6.21%) | 34.00% (±5.88%) | 12.80% (±4.15%) |
| VS-M | 71.60% (±5.60%) | 46.80% (±6.20%) | 30.40% (±5.71%) | 9.20% (±3.59%) |
| ML-DK | 80.00% (±4.97%) | 57.20% (±6.15%) | 41.60% (±6.12%) | 18.40% (±4.81%) |

Table 5. Performance of dynamic komi in handicap games.

Table 5 shows the results for all methods in 1-stone to 4-stone handicap games, denoted by H1 to H4 respectively, against the baseline when playing as white. Generally, both SS-R and ML-DK clearly outperforms all the other methods. ML-DK performs better for H1 and H3, while SS-R performs better for H2 and H4. All in all, considering even and handicapped games altogether, ML-DK appears to perform slightly better than SS-R.

The SS-* methods, regardless of using either rollout or BV for territory estimation, show improvement over not using dynamic komi. The best method seems to be SS-R, while SS-M seems to be the worst among the three SS-* methods. A conjecture for why this is the case is that territory estimates using rollout and BV may be conflicting.

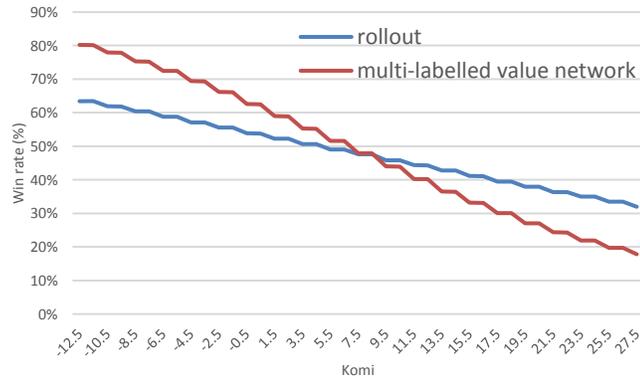

Figure 8. The value of rollout and ML-VN in the opening with different komi.

Second, while the VS method performed the best in [3], in our experiments, the VS-M method did not perform well. In fact, it performs nearly the same as when dynamic komi is not used at all. We observed that the adjustment of dynamic komi using rollouts is relatively slow when compared to that for BV-ML-VN. Figure 8 shows the win rates of rollout and BV-ML-VN with respect to different komi values at the beginning of a game (with an empty board). In the figure, the value of BV-ML-VN is sensitive to each komi value change and is empirically accurate as shown in Subsection IV.C, while the value of rollout is less sensitive as komi changes. Thus, the VS-M method, which only adjusts ±1 each time komi changes, is slow to handle the values obtained from value networks. In [3], since only the rollout was used, it performed relatively well.

In fact, this is what motivated us to design the ML-DK method, which adjusts dynamic komi more responsively. Whenever the score and its corresponding win rate is significantly changed, the next dynamic komi is adjusted quickly and accurately by making full use of the BV-ML-VN.

| Method | H1 | H2 | H3 | H4 |
|---|---|---|---|---|
| SS-R | 75.60% (±5.33%) | 52.00% (±6.21%) | 33.20% (±5.85%) | 10.40% (±3.79%) |
| ML-DK | 73.20% (±5.50%) | 53.20% (±6.20%) | 30.80% (±5.73%) | 10.40% (±3.79%) |

Table 6. Performance of dynamic komi in handicap games when using VN only.

Next, consider the case where only the VN is available. Since multi-labelled values are not available, dynamic methods can only be applied to rollouts. Table 6 shows the results of the two previous best methods when using the SS-R and ML-DK. The results in Table 6 are clearly lower than the same methods in Table 5, showing that having multi-labelled values is clearly superior when they are available.

*E. 6.5-Komi vs. 7.5-Komi*

In this subsection, we perform experiments to demonstrate and analyze the effect of using the value network (BV-ML-VN) trained from 7.5-komi games to play on 6.5-komi games. First, a black-winning game with score 0.5 in a 6.5-komi game is actually a white-winning game with score -0.5 in a 7.5-komi game. Therefore, among all the KGS 6.5-komi games, we collected 2630 games (about 0.68%) ending with score 0.5 for our analysis.

The experiment tries to investigate whether for any of such games, using $v_{6.5}$ from the BV-ML-VN, denoted by VN-6.5, helps in contrast to a value network trained from 7.5-komi games, denoted by VN-7.5. For simplicity, for VN-7.5, we actually use the BV-ML-VN's output $v_{7.5}$, instead of a normal value network trained from 7.5-komi games. To reduce human mistakes, we choose the 10th position from the end of a game as the starting position of a game. For fairness, each game is played twice between the two programs; a program will play as black and white exactly once each for the game. For each move, both programs are given one second.

The experimental result shows a win rate of 51.52±1.22% for the 2630 games. Among the 2630 games, 80 more games are won both times by VN-6.5. This implies that using $v_{6.5}$ does matter, though not by much, which is as expected from the following two reasons. First, in these games, human players may still often miscalculate and make blunders (as described in Subsection IV.C) even for these final 10 moves. Therefore, most of these games may not actually end with score 0.5 for 6.5-komi games. Second, rollout is often sufficient in obtaining a reasonable win rate, which remedies and greatly reduces the problem of using 7.5-komi VN, especially during the end games. Let us illustrate this by a case where black is to lead by 7 points in the optimal play; that is, the position will be winning if the game's komi was set to 6.5 and losing if set to 7.5. Assume black is to move. Let the moves that result in a lead of 7 points be called the optimal moves. In general, the VN based on 7.5-komi games will evaluate that all moves (even for the optimal moves) will have a very low win rate, say nearly 0%. In such a situation, as long as the rollout plays accurately, the mixed value (the average of rollout and VN output) can still differentiate the optimal moves from others. Thus, the program will tend to play the optimal moves. The situation is similar when assuming white is to play, so the discussion is omitted. Hence, most programs find using networks trained from 7.5-komi games to be sufficient.

*F. The Correlation between BV and VN*

In this subsection, the correlation between BV and VN in the BV-ML-VN is depicted in a scatter plot as Figure 9 (below). In this figure, each dot represents a position, and only 2,000 positions are chosen from KGS 7.5-komi games at random for simplicity. For each dot, the y-axis is the value $v_{7.5}$ of the corresponding position, and the x-axis indicates territory based on BV, namely the sum of ownership probabilities of all the board points from the outputs of BV. The figure shows that both are correlated in general, as indicated by a regressed line (in black), centered around (7.5, 50%). Most dots to the right of the vertical line (in red) at 7.5 have values (win rates) higher than 50%, while most dots to the left have values lower than 50%. These dots indicate positive correlation between the board ownership (from BV) and the win rate (from VN). On the other hand, only 6.85% of the dots are in the lower right and 5.3% are in the upper left, which indicate native correlation.

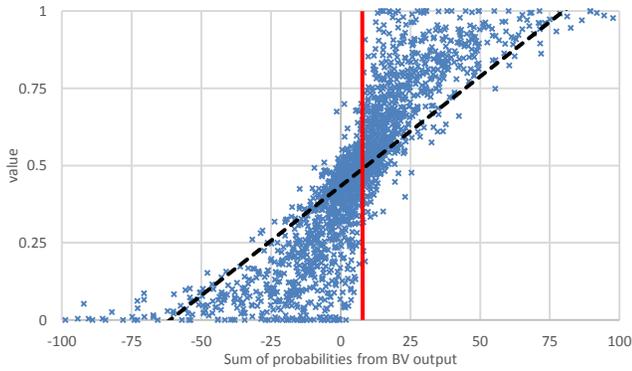

Figure 9. The correlation between BV and VN in 7.5-komi games.

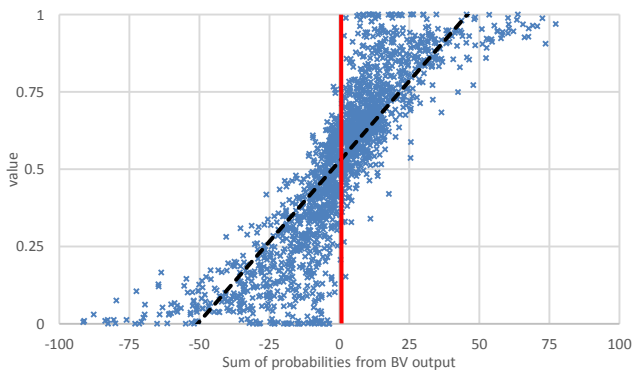

Figure 10. The correlation between BV and VN in 0.5-komi games.

Figure 10 shows another scatter plot on 0.5-komi games. The only difference between Figure 9 and Figure 10 is the value of the y-axis, namely, the y-axis is the value $v_{0.5}$ in Figure 10. Interestingly, we can see all the dots have shifted slightly to the left, and centered around the 0.5 line again (in red). Only 2.7% of the dots are in the lower right and 7.8% are in the upper left, which indicate native correlation.

These figures show that BV gives clues about the win rates of VN. However, the figures do not imply that the values from VN can be directly derived from the ownership probabilities of the BV.

V. CONCLUSIONS

This paper proposes a new approach for a value network architecture in the game of Go, called a multi-labelled (ML) value network. The ML value network has the three advantages, (a) offering different value settings of komi, (b) supporting dynamic komi, and (c) lowering the mean squared error (MSE). In addition, board evaluation (BV) is also incorporated into the ML value network to help slightly improve the game-playing strength.

A metric called the $d$-prediction rate is devised to measure the quality of final score prediction from ML value networks. Experimental results show high confidence on the predicted scores. This shows that the predicted score can be used for dynamic komi. New dynamic komi methods are then designed to support the ML value network, which then improved the game-playing strength of our program, especially for handicap games.

Experiments were conducted to demonstrate and analyze the merits of our architecture and dynamic komi methods.

1. The MSE of the BV-ML value network is generally lower than the VN alone, especially for those games with komi other than 7.5.
2. The program using the BV-ML value network outperforms those using other value networks. Most significantly, it wins against the baseline program, which has just the value network, by a rate of 67.6%. The programs using either just the ML value network or the BV value network also significantly outperform the baseline.
3. For the BV-ML value network, our experiments show the $d$-prediction rates as follows. When $d$ is 0, the $d$-prediction rates are about 8% in the initial stage of the game, growing to 32% near the end. When $d$ increases, the $d$-prediction rates also increase. For example, when $d$ is 6, the prediction rates significantly improve, with about 45% in the initial stage, growing up to 92% near the end. The results show high confidence for short prediction distance, even with the three negative effects described in Subsection IV.C.
4. Experiments show that the proposed ML-DK method improves playing strength. The program with ML-DK wins against a baseline by win rates of 80.0%, 57.2%, 41.6% and 18.4% for 1-stone, 2-stone, 3-stone and 4-stone handicap games respectively. In contrast, the program without using any dynamic komi methods wins by 74.0%, 50.0%, 30.4% and 10.8%. We also investigate other dynamic komi methods such as SS-R, SS-B, SS-M, and VS-M. Among these methods, SS-R performs the best. SS-R also performs better than ML-DK for 2-stone and 4-stone handicap games, but not in any other conditions.
5. Experiments in Subsection IV.E demonstrate that the BV-ML value network, trained from 7.5-komi games, does improve (not by much though) over the value network without ML when playing 6.5-komi games.

6. Our experiments also show high correlation between ML value network output values and BV output values.

From the above, the BV-ML value network proposed by this paper can be easily and flexibly generalized to different komi games. This is especially important for playing handicap games. To our knowledge, up to date, no handicap games have been played openly by programs utilizing value networks. It can be expected that the interest to see handicap games played by programs (e.g., handicap games between top Go programs and top players) will grow, especially when Go programs continually grow stronger. This paper provides programs with a useful approach to playing handicap games.

ACKNOWLEDGMENT

The authors would like to thank the Ministry of Science and Technology of the Republic of China (Taiwan) for financial support of this research under contract numbers MOST 104-2221-E-009-127-MY2, 104-2221-E-009-074-MY2 and 105-2218-E-259-001.

(Submitted to IEEE TCIAIG on May 30, 2017.)

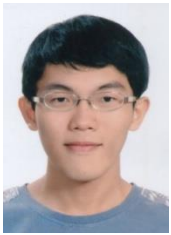
**Ti-Rong Wu** is currently a Ph.D. candidate in the Department of Computer Science at National Chiao Tung University. His research interests include machine learning, deep learning and computer games.

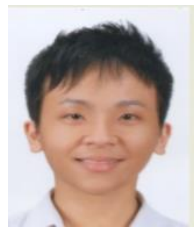
**Hung-Chun Wu** is currently a master student in the Department of Computer Science at National Chiao Tung University. His research interests include deep learning, computer games and grid computing.

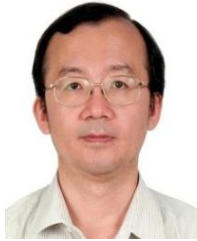
**I-Chen Wu** (M'05-SM'15) is with the Department of Computer Science, at National Chiao Tung University. He received his B.S. in Electronic Engineering from National Taiwan University (NTU), M.S. in Computer Science from NTU, and Ph.D. in Computer Science from Carnegie-Mellon University, in 1982, 1984 and 1993, respectively. He serves in the editorial board of the IEEE Transactions on Computational Intelligence and AI in Games and ICGA Journal. His research interests include computer games, deep learning, reinforcement learning, and volunteer computing.

Dr. Wu introduced the new game, Connect6, a kind of six-in-a-row game. Since then, Connect6 has become a tournament item in Computer Olympiad. He led a team developing various game playing programs, winning over 40 gold medals in international tournaments, including Computer Olympiad. He wrote over 100 papers, and served as chairs and committee in over 30 academic conferences and organizations, including the conference chair of IEEE CIG conference 2015.

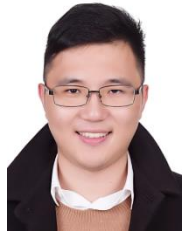
**Li-Cheng Lan** is currently a master student in the Department of Computer Science at National Chiao Tung University. His research interests include machine learning, deep learning and computer games.

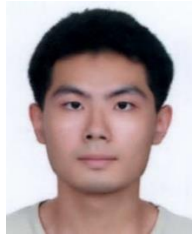
**Guan-Wun Chen** is currently a Ph.D. candidate in the Department of Computer Science at National Chiao Tung University. His research interests include machine learning, deep learning and computer games.

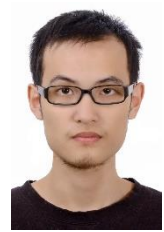
**Ting-han Wei** is currently a Ph.D. candidate in the Department of Computer Science at National Chiao Tung University. His research interests include artificial intelligence, computer games and grid computing.

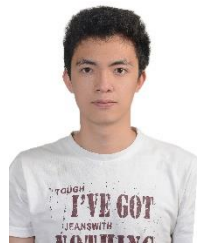
**Tung-Yi Lai** is currently a master student in the Department of Computer Science at National Chiao Tung University. His research interests include machine learning, deep learning and computer games.